# Deep Learning-Derived Optimal Aviation Strategies to Control Pandemics

**Syed Rizvi[1,†], Akash Awasthi[1,†], Maria J. Peláez[2], Zhihui Wang[2,3,4], Vittorio Cristini[2,4,5,6], Hien Van Nguyen[1], Prashant Dogra[2,3,\*]**

[1]Department of Electrical and Computer Engineering, University of Houston, Houston, TX 77004, USA
[2]Mathematics in Medicine Program, Department of Medicine, Houston Methodist Research Institute, Houston, TX 77030, USA
[3]Department of Physiology and Biophysics, Weill Cornell Medical College, New York, NY 10065, USA
[4]Neal Cancer Center, Houston Methodist Research Institute, Houston, TX 77030, USA
[5]Department of Imaging Physics, The University of Texas M.D. Anderson Cancer Center, Houston, TX 77230, USA
[6]Physiology, Biophysics, and Systems Biology Program, Graduate School of Medical Sciences, Weill Cornell Medicine, New York, NY 10065, USA

[†]These authors contributed equally.

\*Correspondence should be addressed to:

**Prashant Dogra, PhD**
Assistant Research Professor of Mathematics in Medicine
Department of Medicine, Houston Methodist Research Institute, Houston, TX, USA
Assistant Professor of Research in Physiology and Biophysics
Department of Physiology and Biophysics, Weill Cornell Medical College, New York, NY, USA
pdogra@houstonmethodist.org

**Keywords:** COVID-19, pandemic, deep learning, graph neural network, artificial intelligence, aviation policy

# ABSTRACT

The COVID-19 pandemic has affected countries across the world, demanding drastic public health policies to mitigate the spread of infection, leading to economic crisis as a collateral damage. In this work, we investigated the impact of human mobility (described via international commercial flights) on COVID-19 infection dynamics at the global scale. For this, we developed a graph neural network-based framework referred to as Dynamic Connectivity GraphSAGE (DCSAGE), which operates over spatiotemporal graphs and is well-suited for dynamically changing adjacency information. To obtain insights on the relative impact of different geographical locations, due to their associated air traffic, on the evolution of the pandemic, we conducted local sensitivity analysis on our model through node perturbation experiments. From our analyses, we identified Western Europe, North America, and Middle East as the leading geographical locations fueling the pandemic, attributed to the enormity of air traffic originating or transiting through these regions. We used these observations to identify tangible air traffic reduction strategies that can have a high impact on controlling the pandemic, with minimal interference to human mobility. Our work provides a robust deep learning-based tool to study global pandemics and is of key relevance to policy makers to take informed decisions regarding air traffic restrictions during future outbreaks.

# 1. INTRODUCTION

In December 2019, the first COVID-19 cases were detected in the Wuhan region of China, from where the SARS-CoV-2 virus rapidly spread worldwide, infecting people and causing the World Health Organization to declare COVID-19 a pandemic in March 2020 (1). As of October 2022, the infection has accounted for over 614 million cases worldwide, with over 6.5 million deaths (2). Since neither vaccines nor therapeutic drugs were available at the onset of the pandemic, governments around the world responded by implementing stringent public health policies to control the spread of infection. These measures included, but were not limited to, social distancing, use of face masks, and travel restrictions. Data analytics and modeling-based studies sought to explore the impact of public health policies on pandemic dynamics and were thus leveraged to optimize the implementation of such policies for maximal impact.

To this end, Kraemer et al. analyzed the effectiveness of public health interventions in China during the early stages of the pandemic, concluding that the interventions significantly mitigated the early spread of COVID-19 (4). Adiga et al. utilized human mobility maps to assess the interplay between mobility data, pandemic dynamics, and public health policy in the United States and India, analyzing potential scenarios of delayed lockdowns and school reopenings (5). International channels of human mobility were also examined to explain early pandemic dynamics; Adiga et al. utilized international air traffic data as a mobility indicator and assessed the effectiveness of flight cancellation policies on the time of arrival of the pandemic to other countries (6).

Further works used mechanistic modeling, machine learning, and deep learning-based approaches to forecast pandemic dynamics and evaluate the impact of public health policies on infection incidence and spread. Kai et al. forecasted the impact of mask mandates on the spread of the pandemic (7), while Anastassopoulou et al. estimated infection spread parameters using a compartmental epidemiological model (8). Chang et al. and Yang et al. integrated population mobility data into epidemiological models to account for the role of population movement in driving the evolution of the pandemic, with the goal of guiding public policy (9,10). A simple seasonal ARIMA model was utilized by Chintalapudi et al. to forecast registered and recovered cases at the onset of lockdown mandates in Italy (11). Ahmar et al. proposed the SutteARIMA model for short-term COVID-19 case forecasting that combined the ARIMA and the α-Sutte indicator (12), and found it to be more suitable for short-term case forecasting than ARIMA (13). Chimmula et al. utilized a deep learning-based approach, applying LSTM (14) networks to forecast Canadian COVID-19 cases and predict the possible stopping points of the pandemic (15). In a novel approach, Dogra et al. leveraged the Elliott Wave principle of financial mathematics to explain and forecast the trends in global COVID-19 cases based on human emotion as a driving factor of the pandemic (3). However, a major limitation of the above works is that they do not incorporate the complexity of spatial relationships and interactions between different geographical locations in determining the outcomes of the pandemic.

To this end, Graph Neural Networks (GNNs) provide a deep learning-based framework that can capture the rich relational information among elements in a network or graph and can thus be leveraged to study the influence of global population mobility on COVID-19 dynamics. Spatiotemporal GNNs are a special class of GNNs that simultaneously consider spatial and temporal information when processing graph inputs, and have been widely applied to problems such as traffic forecasting (16–22). Spatiotemporal modeling with GNNs has also been applied to the problem of pandemic forecasting; Kapoor et al. examined county-level COVID-19 forecasting within the United States by constructing 100 large-scale graph snapshots of US counties, with nodes representing counties and edges representing human mobility between nodes on each day (23). Wang et al. also constructed dynamic mobility graphs using inter-region mobility data at the state-level, and proposed a Recurrent Message Passing (RMP) GNN for mobility-informed infection forecasting (24). Gao et al. developed Spatiotemporal Attention Network (STAN) (25), which involved static edges using both demographic similarity and geographical distance between different locations, and integrated real-world evidence from medical claims into node features to forecast pandemic dynamics. Other works (26,27) constructed models with GNN and LSTM layers to capture both spatial and temporal dependencies in data and predict future cases on European infection data. Sesti et al. devised a GNN-LSTM architecture that operated over a static adjacency graph that was constructed using geographical social connectivity data between different countries (26). Panagopoulos et al., on the other hand, applied a Message-Passing Neural Network (MPNN) to graph snapshots at each timestep of an input window of data, concatenating the output representations and classifying them into future case predictions (27). We observe that these works focus on forecasting COVID-19 cases and pandemic dynamics but lack explainability experiments that could give insight into why a GNN model made a particular prediction. Existing explainability methods for GNNs are defined over static graphs; Pope et al. devised adaptations of common explainability techniques for GNNs, including saliency maps, class activation maps, and excitation backpropagation (28). GNNExplainer (29) introduced a model-agnostic method for finding subgraph explanations of a graph by maximizing mutual information between a graph and its subgraph explanation. These methods, however, are not defined over spatiotemporal graphs, and therefore are limited in their applicability to mobility data that dynamically changes over a temporal dimension.

To address these shortcomings, we developed the Dynamic Connectivity GraphSAGE (DCSAGE) architecture to provide an explainable deep learning-based modeling approach for analyzing mobility-driven regional impact on pandemic dynamics. Our work differs from previous works in that we utilize aviation data as an indicator of international human mobility in the COVID-19 pandemic and focus on the interpretability of our modeling results. In contrast to static-adjacency approaches, we designed our GNN architecture to directly accept dynamic adjacency information that varies on a day-to-day basis. We used sensitivity analysis to quantify the impact of nodes in our spatiotemporal graphs, providing insights into the influence of different geographical regions on pandemic dynamics. From these experiments, we identified the relevance of human mobility (via international flights) in determining the relative impact of various geographical regions, which we quantify as the degree to which a region causes changes in the case predictions of other regions. We use the insights gained on relative impact of different regions to identify tangible strategies for limiting aviation to reduce the impact of highly influential nodes.

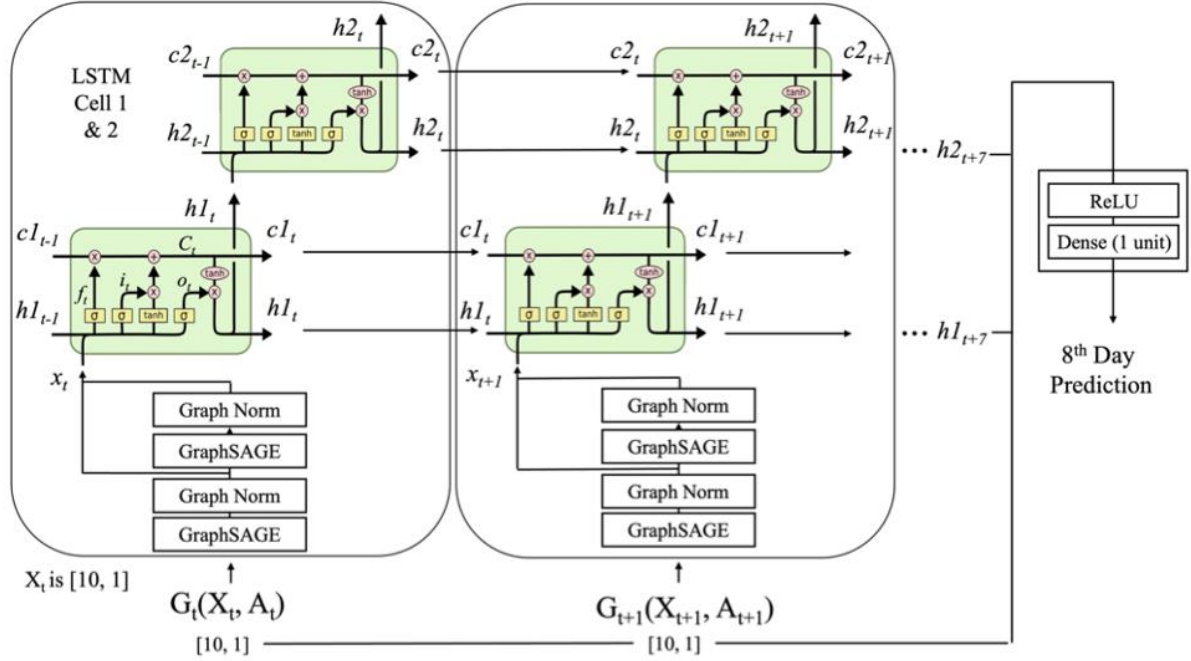

**Fig. 1. DCSAGE model architecture.** A one-day graph $G_t$ is input into the model, consisting of a set of node features $X_t$ (daily COVID-19 cases) and adjacency information $A_t$ (daily flight data). Two sequential GraphSAGE layers process the data to learn spatial representations of the input graph, which is then fed into two stacked LSTM cells that learn spatial relationships over time. After processing seven days of input, the LSTM cell hidden state outputs ($h1_{t+7}$, $h2_{t+7}$) are concatenated with the input node features ($X_t$ to $X_{t+7}$) and passed into the final output layer to predict the next day cases for each node ($\hat{y}$).

## 2. METHODS

### 2.1. DCSAGE model development

To address the challenges of modeling dynamically changing adjacency information in spatiotemporal graphs, we introduced a novel deep learning architecture, referred to as Dynamic Connectivity GraphSAGE (DCSAGE), which utilizes the GraphSAGE message-passing framework (31) to learn spatial relationships between nodes in our graph. As shown in Fig. 1 (see end of manuscript), DCSAGE is a recurrent graph architecture composed of GraphSAGE and LSTM layers, which exploit the spatiotemporal information present in the data. The input for DCSAGE at timestep $t$ is a weighted, directed graph $G_t$, which comprises ten nodes representing the partitioning of the globe into ten geographical locations of interest in the context of the COVID-19 pandemic. The ten nodes are North America, South America, Oceania (i.e., Australia and neighboring island nations), Africa (Egypt excluded), Middle East (includes Egypt), Eastern Europe, Western Europe, Central Asia, South Asia, and Southeast Asia. DCSAGE uses timeseries data for daily COVID-19 infections and international flights for the ten nodes, obtained from public databases (32–34). The infection timeseries comprises each node's feature, while the flight timeseries data is used to weight the edges. The key components of the DCSAGE architecture are described below:

### 2.1.1. GraphSAGE

In contrast to transductive approaches (35), GraphSAGE computes node embeddings in an inductive manner by performing an aggregation operation (such as mean or max) on a sampled set of neighboring nodes and passing the results through a learned nonlinear transformation (31). This allows for new nodes to be introduced into the graph by applying the learned aggregator function on neighboring nodes, opening possibilities for scalability to large graphs and networks.

On a given GraphSAGE layer $k$, GraphSAGE works by sampling a neighborhood $N(v)$ of connected nodes around the current node $v$, and aggregating their embeddings $h_u^{k-1}$ using a permutation-invariant pooling operation $f_{\text{agg}}$, e.g., mean. The aggregated neighbor embedding $h_{N(v)}^k$ is then concatenated with the node's own embedding from the previous layer $h_v^{k-1}$, and fed through a learned embedding function to obtain the node embedding $h_v^k$ for the $k$th layer. The learned embedding function comprises a learned weight matrix $W^k$ and a nonlinear activation function $\sigma$, and the above operations can be represented as: $h_v^k = \sigma(W^k \cdot \text{CONCAT}(h_v^{k-1}, h_{N(v)}^k))$, where, $h_{N(v)}^k = f_{\text{agg}}(\{h_u^{k-1}, \forall u \in N(v)\})$. As a result, the GraphSAGE layer learns to encode the spatial relationship between different nodes.

For the current application of capturing flight-driven pandemic dynamics, we needed to account for both direct and long-distance connecting flights in our DCSAGE model. This is accomplished by using two GraphSAGE layers, allowing for node information to propagate beyond the immediate (or, nearest) neighborhood of a node. Given the limitation of unweighted edges in traditional GraphSAGE, for our application, we modified the message-passing scheme in GraphSAGE by adding flight weights $w_{\text{uv}}$ (weight on the edge from node u to node v) to all edges of the graph, thereby creating a weighted graph for message passing such that the aggregated neighbor embedding is now given by $h_{N(v)}^k = f_{\text{agg}}(\{w_{\text{uv}} \cdot h_u^{k-1}, \forall u \in N(v)\})$.

The GraphSAGE framework includes an optional L2-Normalization step to prevent the magnitude of node embeddings from becoming too large. We omit L2-normalization from DCSAGE in favor of using a separate normalization layer, as described next, since we observed that it improves training convergence.

### 2.1.2. GraphNorm

Internal covariate shift (36) can slow down the training convergence of deep neural networks, due to the changing input distribution for neural layers as the parameters of previous layers are updated. Proposed normalization layers for deep neural networks (36–38) address this problem by shifting and scaling the distribution of activations by its mean and standard deviation and introducing learned parameters to shift and scale output activations, allowing the operation to possibly assume the identity transform. Normalization layers for GNNs have been proposed (39–43), aiming to address the internal covariate shift and oversmoothing challenges in GNNs.

In DCSAGE, we use GraphNorm (39) layers on the output activations of each GraphSAGE layer. GraphNorm is a modification of Instance Normalization (38) which uses a learnable shift rather than subtracting out the total mean statistics of node embeddings, with the intuition that some information may be contained in the statistics of the node embeddings. The normalization operation of GraphNorm on the $j^{\text{th}}$ feature value of node $v$ can be represented as:

$$\text{GraphNorm}(h_{v,j}) = \gamma_j \cdot \frac{h_{v,j} - \alpha_j \cdot \mu_j}{\sigma_j} + \beta_j$$

where $\gamma_j$, $\beta_j$, and $\alpha_j$ are learned parameters, and $\mu_j$ and $\sigma_j$ denote the mean and standard deviation of the feature values across different node embeddings.

### 2.1.3. LSTM

To exploit temporal dependencies across a sequence of one-day graphs, we utilize Long Short-Term Memory (LSTM) cells (14) and learn about spatial relationships in daily infections from previous timepoints. Specifically, we process the output embeddings of the two GraphSAGE layers (containing spatial information) for each daily timestep using two stacked LSTM cells (Fig. 1). LSTMs are a class of recurrent neural networks that control the flow of information using gates. A cell state and hidden state are maintained across the sequence, which capture the spatiotemporal dynamics of the pandemic. At a given timestep $t$, the update within an LSTM can be represented as follows:

$$f_t = \sigma(W_f \cdot [h_{t-1}, x_t] + b_f)$$
$$i_t = \sigma(W_i \cdot [h_{t-1}, x_t] + b_i)$$
$$o_t = \sigma(W_o \cdot [h_{t-1}, x_t] + b_o)$$
$$\hat{C}_t = tanh(W_C \cdot [h_{t-1}, x_t] + b_C)$$
$$C_t = f_t \times C_{t-1} + i_t \times \hat{C}_t$$
$$h_t = o_t \times tanh(C_t)$$

where $f_t$, $i_t$, and $o_t$ are the forget, input, and output gates, respectively; $W_f$, $W_i$, $W_o$, and $W_C$ represent learned weight matrices that are shared across timesteps; $b_f$, $b_i$, $b_o$, and $b_C$ represent bias terms, and $x_t$ is the input of () at a given timestep $t$. $C_t$ and $h_t$ represent the cell and hidden states respectively, and $\hat{C}_t$ is the candidate new cell state formed by processing the input at timestep $t$ along with the hidden state from timestep $t-1$. Parameter updates in recurrent networks are done using the backpropagation through time (BPTT) algorithm (44) or truncated BPTT (45) in the case of long sequences where information in error gradients would become diluted.

### 2.1.4. Skip Connections

After multiple iterations of message passing, GNNs often suffer from a problem known as oversmoothing (46–48), where the embeddings for different nodes become too similar. This occurs when the receptive fields (K-hop neighborhood) of different nodes become highly overlapped (49). To prevent oversmoothing in DCSAGE, we make two design choices: (i) we set the number of GNN layers equal to the most common graph diameter (two) found on daily input graphs in our dataset (50), and (ii) introduce a concatenation-based skip connection (51) from the first GraphSAGE layer output to the LSTM, so that both one-hop and two-hop information are passed to the LSTM Cells. Including a skip connection here effectively results in learning a mixture of one and two-hop models (52,53), and has been successfully applied in other works to prevent over smoothing in GNNs (23,54).

We reuse the input node features by adding a second skip connection from the input cases to the final linear output layer, shown in Fig. 1. Input node features $X_t \ldots X_{t+7}$ are concatenated with the output of the two LSTM cells ($h1_{t+7}$ and $h2_{t+7}$) at the last timestep and used to predict the next-day cases. This skip connection ensures that input information, which may be changed or washed out as it passes through the network (51), reaches the linear output layer. Additionally, it alleviates the distribution shift between LSTM cell outputs which are in range [-1, 1] and COVID-19 cases (>0), which need to be non-negative numbers and can span across several orders of magnitude.

### 2.1.5. Output Layer

In order to make predictions at timestep *t*+8, we first concatenated the output hidden states of both LSTM cells at timestep *t*+7 with the input features from the above-mentioned input feature skip connection (from days *t* to *t*+7) and apply the rectified linear unit (ReLU) function (55) to ensure that models output are non-negative numbers. We then apply a linear layer with 1 neuron to classify the embedding for each node into a case prediction. This can be represented as:

$$x = ReLU(concat(h1_{t+7}, h2_{t+7}), X_t, X_{t+1}, \ldots, X_{t+7})$$
$$\hat{y} = Wx + b$$

where $h1_{t+7}$ and $h2_{t+7}$ are the hidden outputs of LSTM cell 1 and 2 at timestep $t=7$, $X_t$, ..., $X_{t+7}$ are the input node features for each timestep, and $W$ is a learned weight matrix.

### 2.2. Data curation

To train our model, we used data from public databases involving daily COVID-19 infections and international flights. For the node feature timeseries data, we summed-up the daily infection numbers across various countries belonging to a given node (Fig. S1); to obtain the dynamic, bidirectional edge weights, we aggregated daily international flights departing from or arriving to the various airports across a country (Fig. S2) and summed the numbers across countries in a given node (Fig. S3).

For data pre-processing, we first aligned the timespan of both data sources and deleted any days where information was missing or erroneous (negative values), resulting in a timeframe from March 1st, 2020, to September 30th, 2021, with intermittent gaps. We then smoothened the infection data using a 7-day moving average, after which we performed a $log_{10}$ transformation to reduce the skewness in the dataset (Fig. S4). The flight data was not smoothened but was $log_{10}$ transformed. A 64%/16%/20% split was used for the training/validation/test dataset, respectively, with the validation and test data lying subsequent to the training data in the timeseries.

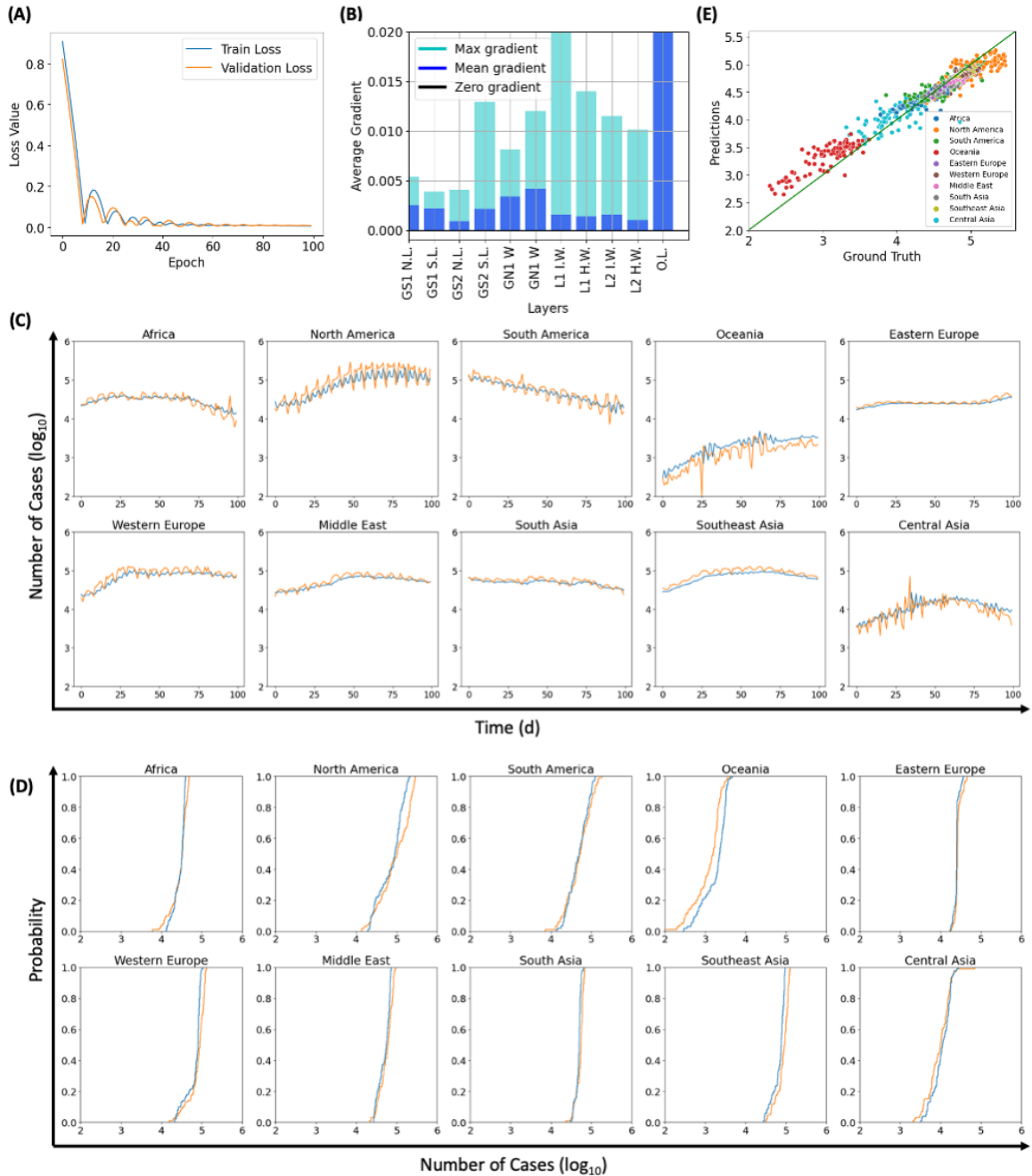

**Fig. 2: Model Training and Predictions.** (A) Training and validation loss curves for DCSAGE. (B) Gradient flow plot taken from epoch 10 during training of a representative DCSAGE model. GS1 and GS2 indicates GraphSAGE layers 1 and 2 respectively, with N.L. indicating the neighbor linear layer which processes the aggregated neighbor embedding and S.L. indicating the self linear layer which processes the node's own representation. GN1 W and GN2 W indicates the weight of GraphNorm layers 1 and 2, respectively, while L1 and L2 denotes LSTM Cells 1 and 2, with I.W. indicating weights which process inputs to the LSTM cell and H.W. indicating weights which processes the hidden state from previous timesteps. O.L. denotes the output linear layer which outputs 8[th] day case predictions for each node. (C) Prediction trends over a period of 100 days comprising our test dataset, with each daily prediction made based on a 7-day input window of test data. (D) Cumulative Distribution Function plots of DCSAGE predictions in blue versus ground truth cases in orange. (E) Scatterplot of ground truth versus DCSAGE predictions, color coded by node.

### 2.3. Model training

For model training, we give the model a 7-day input window of data (hereafter referred to as a data input window) from days t to t+7 and predict the number of cases ŷ for each node at timestep t + 8. We define our loss function, L, as the Mean Absolute Scaled Error (MASE) between prediction ŷ and ground truth cases $y$, such that

$$L = \frac{\frac{1}{B}(\Sigma_{i=0}^{W} \hat{y} - y)}{\Sigma_{i=0}^{W} y},$$

where $W$ denotes the number of data input windows in a single batch, and $B$ denotes the number of batches in our dataset. Note that we set $B$ equal to 1, i.e., we fit our entire dataset into memory at once and update learnable parameters in our models using batch gradient descent (56), represented as:

$$\theta_{t+1} = \theta_t - \eta \cdot \nabla L,$$

where $\theta_i$ represents learned parameters at time $i$, $\eta$ is the learning rate, and $\nabla L$ is the gradient of the loss function computed across the entire dataset. Calculating gradients across all training samples is possible due to the small size of our dataset, and results in more accurate parameter updates. We train for 100 epochs to achieve convergence, saving the model that yields the lowest value of loss on the validation set. To update model parameters, we use the Adam optimizer (57) with an initial learning rate of 1e-2 and implement learning rate decay during the training of DCSAGE with a patience of 40 epochs. Adaptive optimizers such as Adam use per-parameter learning rates to adaptively optimize different parameters in models with different learning rates, allowing for faster training convergence.

Weight initialization in DCSAGE is done with Kaiming uniform initialization (58) for linear layers within GraphSAGE and uniform initialization within LSTM layers. Graph Neural Networks can lose expressive power when their capacity, defined as their depth · width (59), is restricted. Given our choice of a 2-layer GNN, we set the embedding dimension (width) of our GraphSAGE layers to 10 and empirically find that it is sufficient for our forecasting task. We implement all models in PyTorch (60), a general purpose deep learning library in Python.

### 2.4. Model predictions

To assess the impact of different aviation strategies on global case trends, we must be able to forecast COVID-19 cases over a period of time into the future. With this ability, we can run perturbation experiments on model inputs , analogous to real-life aviation restrictions, and assess the impact on predicted future cases. There are several strategies for long-term time series forecasting; a direct way would be to train a single many-to-many architecture to accept a window of input data and predict the necessary window of future cases. This, however, leads to problems with accurate predictions due to the difficulty in optimizing an entire window of future predictions. Thus, other works (26,27) instead train a many-to-one architecture to predict one timestep of data based on a window of input. This gives a more accurate prediction for some single future timestep $t + d$, where $d$ denotes the lead time after the last input at timestep $t$. Multiple many-to-one models can then be utilized to form a long-term forecast, with each model focusing on predicting a specific lead time into the future (27). This approach, however, requires training a separate model for each prediction timestep desired, which is computationally burdensome. We instead chose to train a many-to-one architecture and obtain a long-term forecast of length L by iteratively predicting one day into the future L times while feeding model predictions back to itself to make the input for predicting future timesteps (recursive prediction). We define the long-term forecast of length L made in this fashion as a recursive prediction window. The drawback of this strategy is that model prediction error will be accumulated with each recursive iteration.

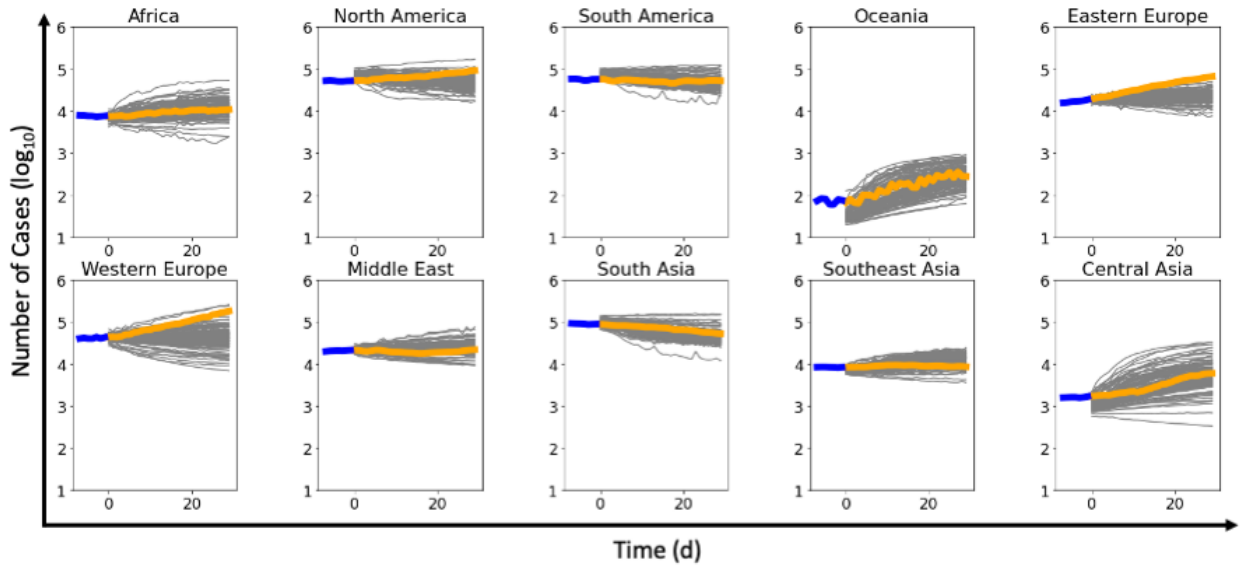

**Fig. 3: Bias Correction.** Recursive predictions of 100 DCSAGE models on dataset window 200 after bias correction, with 1 scaling factor being applied to each node.

### 2.5. Bias Correction

To address the prediction biases within our model, we correct all model predictions using correction (or scaling) factors. Given D days of recursive predictions made on W 7-day data input windows within the dataset, the bias correction factor $b_c$ for a single geographical region c is calculated as follows:

$$b_c = \frac{1}{W}\Sigma_{w=1}^{W}\frac{1}{D}\Sigma_{d=1}^{D}\frac{y}{\tilde{y}}$$

where $\tilde{y}$ is the averaged prediction of 100 DCSAGE models ($= \frac{1}{100}\Sigma \hat{y}$) and y is the ground truth for that data input window. In practice, we have W = 523 data input windows in our dataset, and we obtain D = 30 days of recursive predictions from our models on each data input window. This yields a single correction factor $b_c$ for each of the 10 nodes in the graph, which we then utilize to correct further model predictions by multiplying the prediction by its correction factor. We provide calculated bias correction factors in Table S1.

## 2.6. Local Sensitivity Analysis

We first obtained 30 days of recursive predictions (recursive prediction window of length 30) for all (W = 523) data input windows in the dataset, referred to hereafter as unperturbed recursive predictions. Perturbed recursive predictions were then obtained by removing all incoming and outgoing edges from a chosen node, one node at a time, and again obtaining 30 days of recursive predictions, yielding 10 sets of perturbed recursive predictions corresponding to the 10 perturbed nodes in our graph. This effectively isolates or "locks down" the node, analogous to restricting flights to and from a geographical location. We can define the impact that a geographical region has on global cases - hereafter referred to as its sensitivity - as the absolute value of the difference in model predictions summed over 30 days of recursive predictions and across the other 9 nodes affected by the perturbation. On a single 7-day data input window given to the model to start recursive prediction, this sensitivity measure can be represented as follows:

$$\text{Node sensitivity} = \sum_{n=1}^{N-1} \sum_{d=1}^{D} |f(G) - f(G \setminus G_s)|$$

where $N$ is the number of nodes in the graph, $D$ is the number of days of recursive prediction from the model, f represents the model, and $G_s$ is a subgraph of graph $G$. In all our experiments we set $N = 10$ and $D = 30$. We note the similarity of this spatiotemporal sensitivity measure to the fidelity metric proposed in (28):

$$\text{Fidelity} = \frac{1}{T} \sum f(G) - f(G \setminus G_s)$$

where $D$ is the number of graphs in the test set. Intuitively, the fidelity metric quantifies how discriminative an identified subgraph is to model predictions. Our metric differs from fidelity in that we extend our metric to spatiotemporal graphs, and in practice our removed subgraph is always a single node for which we are trying to quantify the global impact.

## 2.7. Policy impact quantification

We quantified the overall impact of a policy by first summing the absolute difference between perturbed and unperturbed predicted global cases over 30 days for each recursive prediction window and via each model. This summed difference in global cases was averaged across 40 models and then subsequently averaged across all 523 recursive prediction windows, yielding a single number. This calculation can be represented as follows:

$$\text{Policy Impact} = \frac{1}{W} \sum_{w=1}^{W} \frac{1}{M} \sum_{m=1}^{M} \sum_{d=1}^{D} |\tilde{y} - y|$$

where $\tilde{y}$ is the perturbed global number of cases, $y$ is the unperturbed global number of cases, $W$ is the number of recursive prediction windows, $M$ is the number of models used, and $D$ is the number of days of recursive prediction. For our experiment we use $W = 523$, $M = 40$, and $D = 30$. The policy impact was normalized by the value corresponding to the most impactful policy, thus providing a dimensionless numbers indicating relative policy impact.

## 3. RESULTS

### 3.1. Model Training, Predictions, and Bias Correction

Our model aims to learn the dynamic spatiotemporal patterns in the global evolution of the COVID-19 pandemic, based on human mobility assessed through international air traffic data. With an input of 7 graphs representing daily information of COVID-19 cases per node and international flights per edge, the model predicts new infections for each node on the $8^{th}$ day. Deep learning models exhibit stochasticity in their predictions due to differences in parameter initialization and training, therefore we train 100 DCSAGE models and consider average predictions for our numerical experiments and analyses. As shown in Fig. 2A, loss curves from the training of a representative DCSAGE model show convergence within 100 epochs, thereby minimizing the error between model predictions and ground truth. The tight coupling observed between the train and validation loss shows that the model generalizes well to new data, predicting COVID-19 cases accurately on validation data which comes chronologically later in the dataset than the training data.

The regularization in DCSAGE, which comes from ReLU activations and GraphNorm layers, contributes to limiting model capacity and preventing overfitting on the training dataset. We plot the gradient flow in DCSAGE during training in Fig. 2B, which shows the magnitudes of updates being applied to each layer within DCSAGE at a given epoch during training. Extremely high or low weight updates (denoted by fully cyan bars and the lack of any bar, respectively) are indicative of exploding or vanishing gradients, both of which harm training. The plot shows that gradient updates flow smoothly back to the input GraphSAGE layers in DCSAGE, indicative that they are receiving weight updates and learning spatial relationships and thus pandemic dynamics between different geographical regions.

As shown in Fig. 2C, the representative model predicts $8^{th}$ day cases in close agreement with the ground truth across the entire testing dataset of 108 days. This is also evident in the cumulative distribution function plots (Fig. 2D) and Pearson correlation analysis (Fig. 2E). The Pearson correlation coefficient $R = 0.97$ indicates high degree of correlation between model predictions and ground truth. Note that similar to training, each $8^{th}$ day prediction in the test set evaluation was based on one 7-day input of test data.

Using the 100 trained models, we obtained 30 days of recursive case predictions from a 7-day data input window. As shown in Fig. 3, the recursive predictions for most nodes deviate from the corresponding ground truth, most likely due to accumulation of systemic errors in subsequent daily predictions, for instance small overpredictions and underpredictions observed for Oceania and North America, respectively (Fig. 2C,D). Therefore, to correct for such deviation, we estimated a correction factor for each node by taking the ratio of ground truth to average model predictions (see Methods) and applied it to the original predictions. As shown in Fig. 3, the ground truth trend is mostly centered upon the overlaid band of corrected recursive predictions, thereby allowing us to reliably use the recursive prediction strategy for further analyses.

We use MPNN (27) as described by Panagopoulos et al. as a reference for comparing the performance of the DCSAGE model, and provide corresponding training figures for MPNN in Fig. S8. MPNN consists of LSTM (16) and GCN (37) layers which process graph inputs representing a week of input data; each node contains the previous 7 days of COVID-19 cases as features, and an averaged graph structure over the 7 input days is input into MPNN as the adjacency information. We note that this is not as flexible as the dynamic modeling of adjacency connections in DCSAGE, which can change by day to better model human mobility indicators. As shown in Fig. S9, MPNN models exhibit more variability in predictions over a 30-day recursive window, making sensitivity score calculations based on recursive predictions less reliable. For this reason, we do all further analysis using DCSAGE.

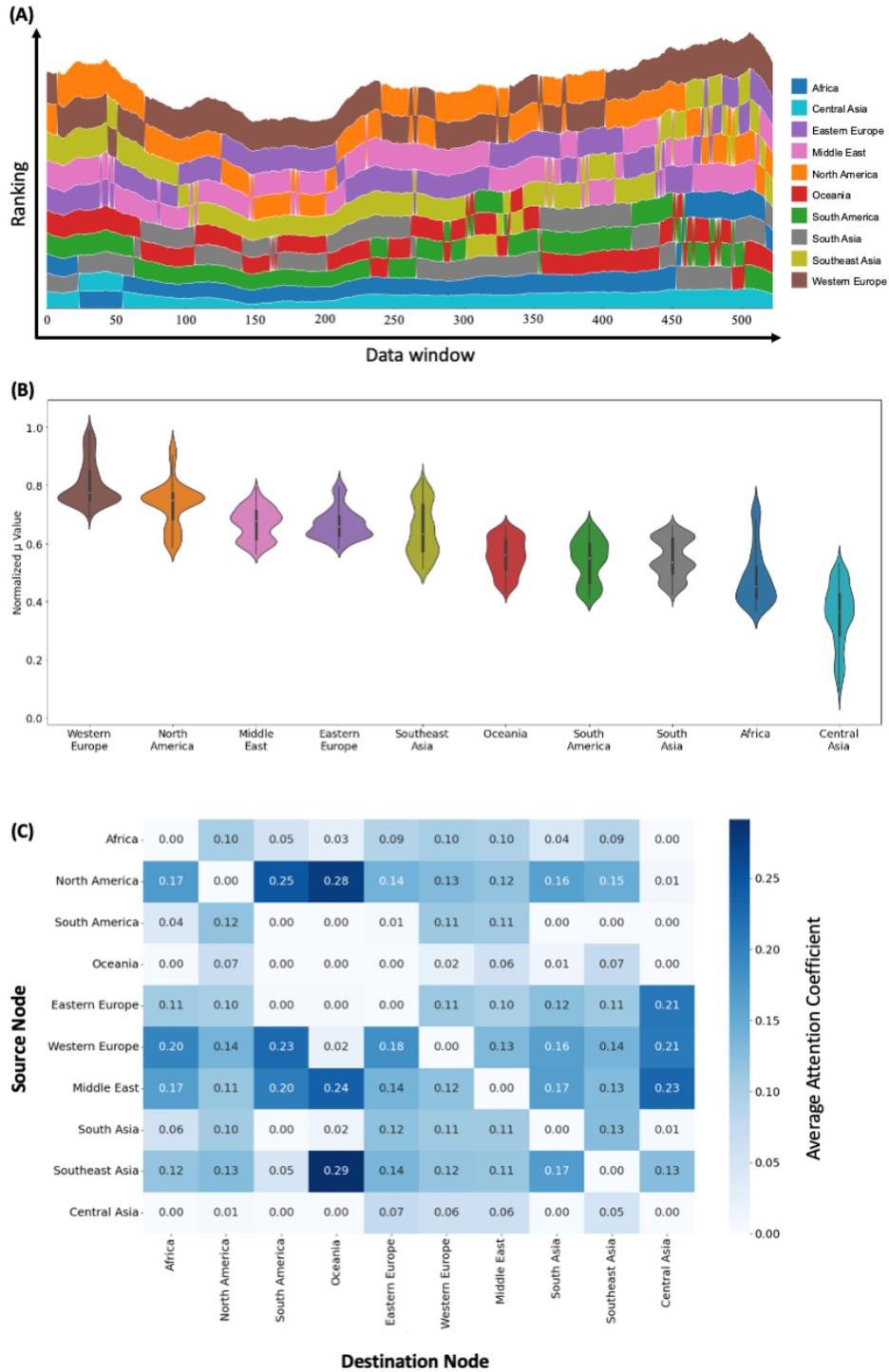

**Fig. 4: Sensitivity Analysis.** (A) Sensitivity score rankings for the 10 geographical regions over all windows in the dataset based on fitted EV distribution mu parameter, depicted as an area bump chart where the height of the series for a geographical region corresponds to the magnitude of its mu parameter. (B) Violin plot of mu parameter distributions for each of the 10 regions, arranged in order of distribution median. Each distribution is comprised of the mu parameters taken across all windows in the dataset. (C) Attention coefficient matrix produced from node-level attention coefficients of the first Graph Attention layer in DCGAT. Each cell represents attention coefficients for a particular edge connecting two regions (if present) summed across all one-day graphs in the dataset, with rows being the source and columns being the destination region of the edge.

## 3.2. Sensitivity Analysis with Spatiotemporal Node Perturbation

Interpretability in deep neural networks can be defined as the degree to which an observer understands the decision of a model (61). This can be broken down into interpretable modeling techniques, which aim to increase the transparency of the model itself, and post-hoc explainability methods. To obtain explainable predictions from our DCSAGE model, we need explainability methods that are well-defined over spatiotemporal graphs. Existing explainability methods, however, provide post-hoc explanations on static graphs, and do not extend to the case of spatiotemporal graphs with dynamically changing adjacency information. To address this shortcoming in spatiotemporal explainability methods, *we define a novel perturbation framework on spatiotemporal graphs which removes edges within the graph to isolate nodes and quantify the effect on model predictions*.

Using the forecasting ability of our model, we predict future cases based on aviation trends and employ it to identify the impact of various nodes on global evolution of the pandemic. For this, we used sensitivity analysis where a "perturbation" of a single node in the network is imposed to quantify the effect on other nodes in terms of change in daily cases. We used the predictions from the previously trained 100 models to calculate a sensitivity score for each node across the duration of study (see Methods), with higher scores indicating a larger impact on the development of the global pandemic. Note that considering calculations from 100 models allows us to account for the inherent stochasticity of deep neural network models, which can be minimized by averaging the results. We choose to analyze pandemic dynamics through regional sensitivity rather than directly forecasting cases under perturbed settings due to the limitations of our model in direct case forecasting scenarios (see Limitations).

As shown in Fig. S5, the sensitivity score timeseries from 100 models indicates that more models predict higher sensitivity scores for North America and Western Europe, suggesting greater impact of these nodes over the pandemic. We verified this by performing increasingly large flight reductions on North America as a representative node, which yielded higher sensitivity scores as more flights were reduced (Fig. S6). To quantify the average sensitivity score on each recursive prediction window for each node, we fit a Gumbel distribution (right-tailed distribution) to the predicted sensitivity scores of the 100 models on each recursive prediction window and estimated the $\mu$ parameter, which represents the mode of the distribution (Fig. S7). The $\mu$ value is then used to rank order the nodes in terms of their sensitivity. As shown in the area bump chart in Fig. 4A, for a major portion of the studied timeframe, Western Europe ranked the highest, closely followed by North America, with the two nodes frequently switching places. The rankings of Middle East, Eastern Europe, and Southeast-East Asia fluctuated between $3^{rd}$ and $5^{th}$ places during the pandemic, succeeded by Oceania, South America, and South Asia. Africa and Central Asia were at the bottom of the ranking list, suggesting low impact of these two nodes on the evolution of the pandemic. The timeseries results are summarized as violin plots in Fig. 4B, which rank ordered the distribution of $\mu$ values across the timeseries based on the median of the distribution, and the observation are consistent with the trend observed ii the timeseries plot ($\mu$ values normalized between 0 and 1).

Finally, to validate the observations made from sensitivity analysis, we introduced interpretability into our model in an alternative manner using attention mechanisms, which calculated attention coefficients that can be interpreted as importance scores contributing to the predictions of the model (62,63). We implemented this verification in the DCGAT variant of our model, which replaces the GraphSAGE layers in DCSAGE with Graph Attention layers (30). This provides attention scores on the directed edges of our dynamic one-day graphs while also preserving the inductive property of our model, allowing us to investigate which geographical regions are being given more importance in model predictions. We calculated attention coefficients with a representative DCGAT model over all days in our dataset post-training and aggregated them into an attention coefficient matrix representing all possible edges in our graph. As shown in Fig. 4C, the cumulative attention score tends to show higher values for edges originating from Western Europe, Middle East, and North America, consistent with our previous findings (Fig. 4B). Similarly, the remaining nodes follow a similar trend as observed through sensitivity analysis with Oceania, Africa, and Central Asia exhibiting the lowest scores.

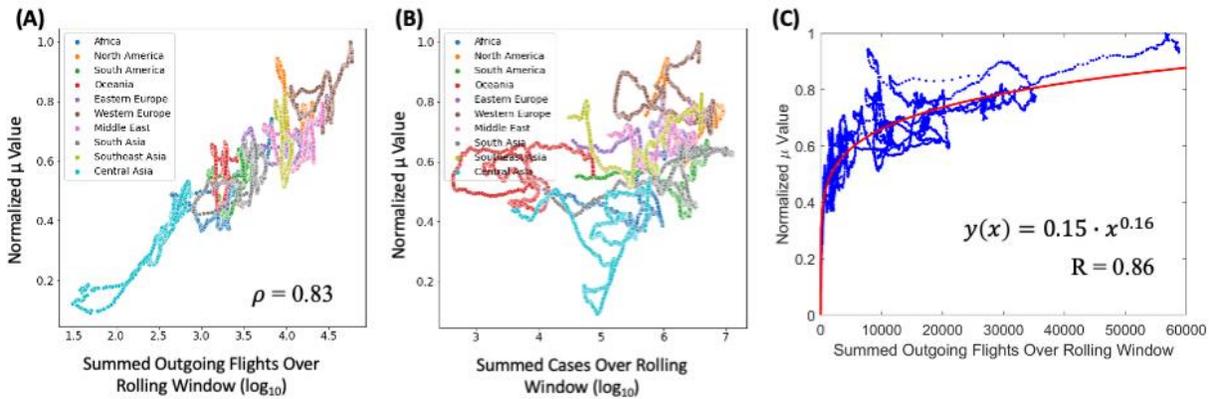

**Fig. 5: Flight and COVID-19 Cases Plotted Against Sensitivity Scores**. (A) Outgoing flights of different geographical regions plotted against corresponding sensitivity mu parameters calculated during spatiotemporal sensitivity analysis. Rho ($\rho$) indicates spearman correlation calculated between flight numbers and mu parameters. (B) Summed COVID-19 cases per window for each geographical region versus corresponding mu parameters. (C) Fitted curve on the flights versus mu parameter scatterplot, following a power law relationship with a strong correlation (R = 0.86) between fitted curve values and mu parameter values.

### 3.3. Outgoing flights as a correlate of node impact

To understand the relative impact of human mobility and infection load on the global evolution of the pandemic, we evaluated the correlation between mobility and daily cases versus model-derived sensitivity scores. As shown in Fig. 5A, nodes with larger number of outgoing flights tend to have a higher sensitivity score, while that is not necessarily true for the daily case count (Fig. 5B). For instance, Oceania (red) and South America (green) despite a huge disparity between their case count distributions over 567 days (Fig. 5B), have comparable sensitivity scores which can be attributed to their similar outgoing flight counts (Fig. 5A). Alternatively, Central Asia with larger number of daily cases than Oceania has a smaller sensitivity score due to its lesser number of outbound flights compared to Oceania. Similarly, South Asia has one of the largest numbers of daily COVID-19 cases at certain timepoints, however it is still superseded in impact by North America, Europe, and the Middle East due to the higher number of outgoing flights emanating from those regions.

With these observations, we can establish outgoing flights as a correlate of node impact or sensitivity. This is further corroborated by a strong Spearman correlation between outgoing flights and sensitivity scores for all the ten nodes pooled together (Spearman's rank correlation coefficient $\rho = 0.83$; Fig. 5A). This nonlinear correlation is characterized by a power law that is in good agreement with the raw data (Pearson correlation coefficient $R = 0.86$; Fig. 5C). It can thus be inferred that aviation trends hold the key to controlling the global evolution of the pandemic, and attention must be given to geographical regions that are aviation hubs, irrespective of the magnitude of their infection load. With the ability of our dynamic network model to use real-time human mobility data and identify the most influential geographical regions driving the pandemic, we can propose adjustments in policies relevant to international flights necessary to control the pandemic.

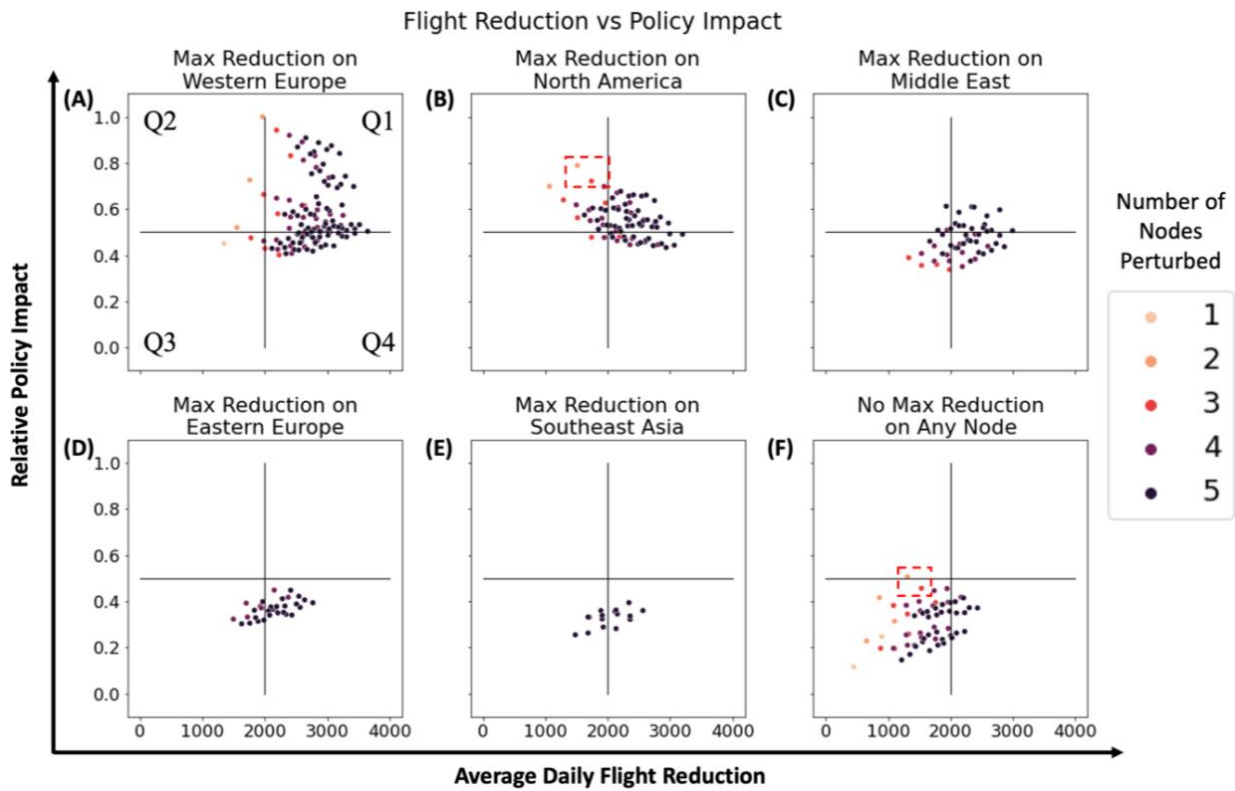

**Fig. 6: Policy Implementation.** (A) Scatter plot of average daily flight reduction versus policy impact policies which included a 75% reduction on incoming and outgoing flights to Western Europe. The subplot is divided into four quadrants, with Q1 indicating high flight reduction and high impact, Q2 denoting low reduction and high impact, Q3 denoting low reduction and low impact, and Q4 indicating high reduction and low impact. (B) Scatter plot of average daily flight reductions versus policy impact of policies which included a 75% reduction on North America. The red dotted box indicated a few policies of interest in Q2, which have high relative impact with less flight reductions compared to other policies. (C) Scatter plot of policies including a 75% reduction in flights to and from the Middle East. (D) Scatter plot of policies including a 75% reduction on Eastern Europe. (E) Scatter plot of policies including a 75% reduction on Southeast Asia. (F) Scatter plot of policies that do not place a 75% reduction on any of the five most sensitive nodes in the graph. Two policies of interest are demarcated by the red dotted box in Q2, highlighting policies that have higher impact while not making 75% flight reductions on any node.

### 3.4. Forecasting flight reduction effects on pandemic control

Based on our findings from the local sensitivity analysis (Fig. 4), we explored creating sets of flight reductions which, when applied to highly sensitivity nodes, would be impactful in terms of relative impact on all nodes. As a tangible proof-of-concept, we tested policies (363 total) that involved reductions in incoming and outgoing flights at different levels (25%, 50%, and 75%) in up to the five most sensitive nodes (Western Europe, North America, Middle East, Eastern Europe, and Southeast Asia). 40 out of the 100 previously trained and randomly selected models were used for this analysis (see Methods). As shown in Fig. 6, policy outcomes can be divided into four quadrants based on the extent of reduction in flights and the corresponding impact achieved, i.e., (Q1) high (flight) reduction, high impact, (Q2) low reduction, high impact, (Q3) low reduction, low impact, and (Q4) high reduction, low impact. Also, in Fig. 6 we present the results such that policies with 75% reduction (i.e., the max possible reduction in our analysis) on a given node are plotted together. Each policy may have reductions in multiple nodes, where the number of perturbed nodes is indicated by the color coding in each graph. Note that policies involving 75% reduction in multiple nodes are only presented once, giving priority to the graph with a more sensitive node.

As shown in Fig. 6A, when a 75% reduction is applied to at least Western Europe, with or without reduction in the other four nodes, the outcomes mostly end up in Q1. This indicates that a high impact on pandemic control can be achieved if policies include stringent restriction on highly sensitive nodes. Of note, two policies are identified with high impact (demarcated in dotted square) while only reducing around 2000 daily flights on average. Stringent flight reductions on North America (with other node reductions combined) also yield high impact on the pandemic, i.e., mostly Q1 and Q2-type response (Fig. 6B). Two policies are again highlighted that impose a reduction of around 2000 flights, however these have lower relative impact than had the large restriction been put on Western Europe. Through this we identified that simultaneously perturbing the most sensitive nodes (e.g., 75% Western Europe, 50% North America) is a more strategic approach than perturbing multiple, less sensitive nodes. This is evidenced by the results shown in Fig. 6C-E where policies with 75% reduction in the lesser sensitive nodes (Middle East, Eastern Europe, and Southeast Asia) led to lower impact for similar or larger flight reductions. On Fig. 6F, we note that with no max reductions on highly sensitive nodes, policies imposing lesser reductions resulted in lower impact (i.e. Q3 outcomes) even when the average daily flight reduction matched that of more impactful policies.

### CONCLUSIONS

In this work, we investigated the relationship between human mobility in the form of international flights and global COVID-19 pandemic dynamics. We developed a novel graph neural network-based model known as DCSAGE, capable of processing one-day graph inputs with dynamically changing adjacency information. A perturbation methodology was designed on our spatiotemporal GNN model to identify the impact of geographical regions on global pandemic dynamics, from which we identified that Western Europe, North America, and Middle East have the highest impact on fueling the pandemic, which correlates with the larger volumes of outgoing international flights. We use these insights to set up and search through tangible restriction policies on air traffic for controlling the pandemic, identifying policies and strategies with large impact at lesser air traffic reductions. Our work represents a novel usage of perturbation analysis on spatiotemporal GNNs to gain insight on pandemic forecasting. Our method, however, can be extended on by integrating more node and edge features into the model, for example by including socioeconomic or development factors into the features of nodes in our graph, or including additional edge features such as geographical proximity and political relationship in the event of future outbreaks.

Under perturbation scenarios where flights are removed for geographical regions to analyze sensitivity, the change in forecasted cases from DCSAGE models is not constrained to be negative. Since models are not guaranteed to forecast reductions in cases when flights are perturbed, we choose to quantify the sensitivity of a node by observing the absolute value of the change in forecasted COVID cases in the rest of the nodes. We then sum up the predicted change on the rest of the graph to get the relative impact of a node on the rest of the graph. Due to this limitation, we focus our analysis around the relative impact of geographical regions.

## Acknowledgements
This work was supported in part by the Cockrell Foundation (PD, VC), the National Institutes of Health (NIH) grants 1R01CA253865 (ZW, VC), 1R01CA226537 (ZW, VC), and 1R01CA222007 (ZW, VC). The funders had no role in study design, data collection and analysis, decision to publish, or preparation of the manuscript. PD also acknowledges Dr. Vrushaly K. Shinglot and Carmine Schiavone for helpful scientific discussions.

## Competing Interests
The authors declare that no competing interests exist.

# SUPPLEMENTARY INFORMATION

## I. Dataset Figures

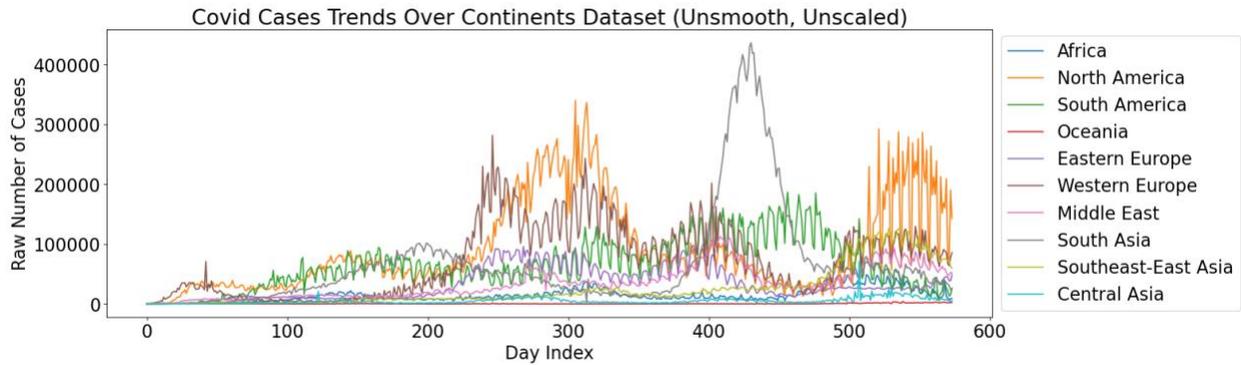

**Figure S1. COVID-19 cases per day.** Daily COVID-19 infections trend for each geographical region plotted across the dataset.

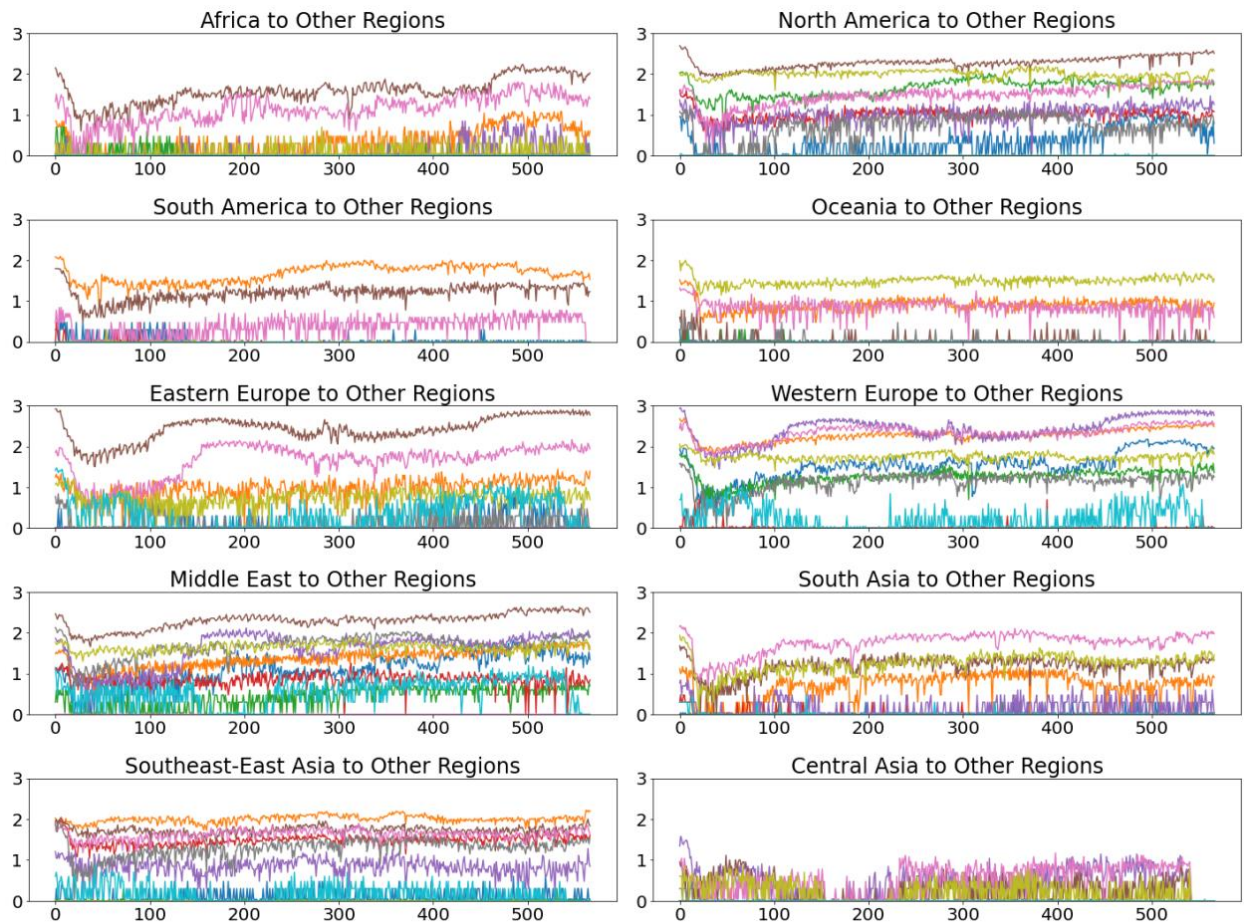

**Figure S2. Outgoing flights per geographical region.** Outgoing international flights plotted for each geographical region across the dataset. Each subplot depicts each of the 9 other destination regions color coded.

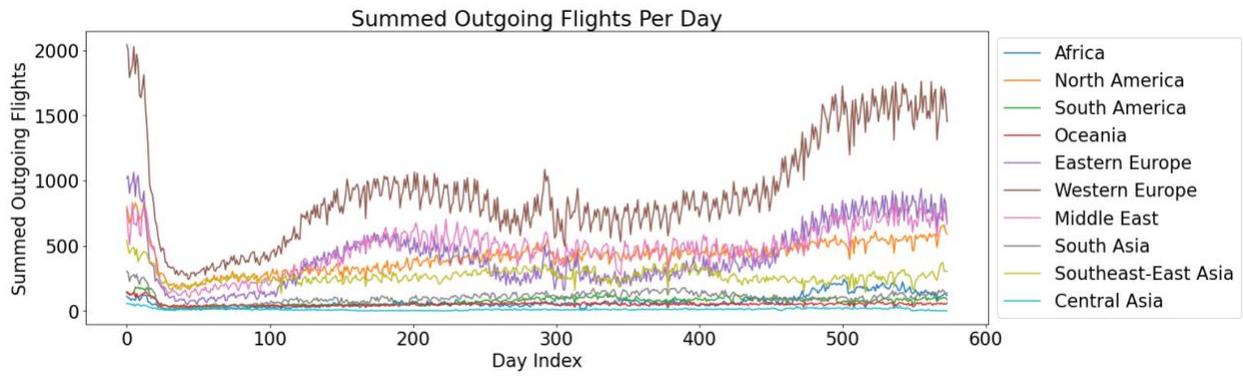

**Figure S3. Total outgoing flights per day.** Outgoing total international flights of each geographical region plotted across the dataset.

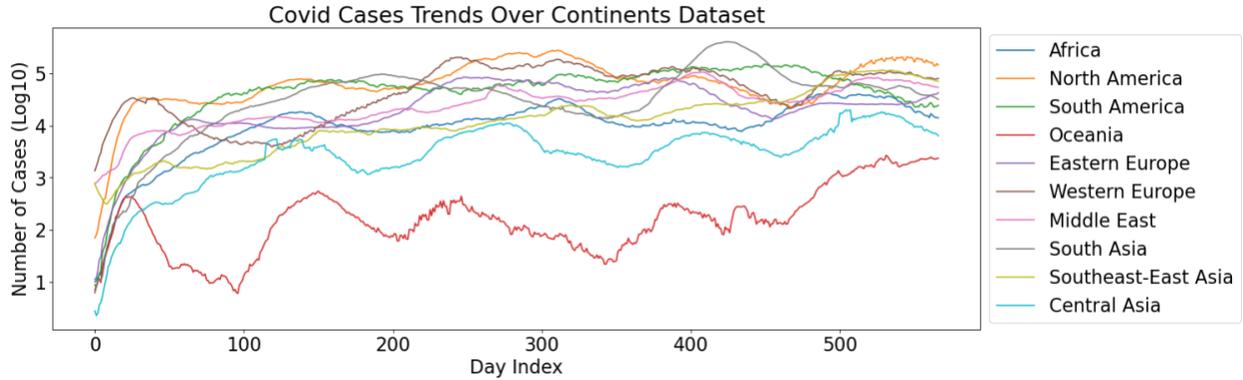

**Figure S4. Smoothened and log$_{10}$ transformed COVID-19 cases.** Daily COVID-19 infections trend for each geographical region after smoothing with a 7-day moving average and log$_{10}$ transformation.

## II. Additional DCSAGE Results

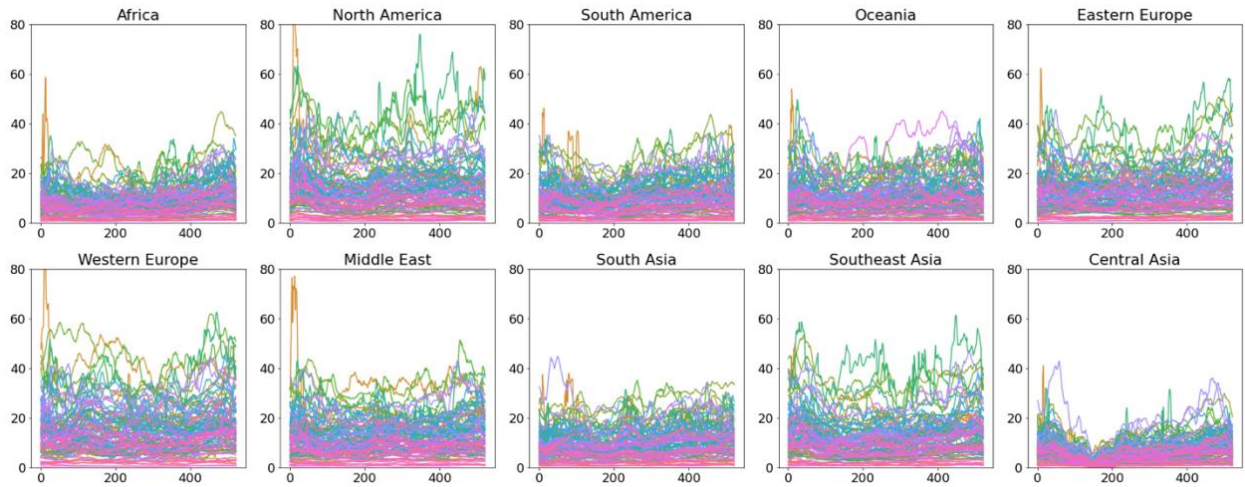

**Figure S5. DCSAGE sensitivity score trends.** Sensitivity score trends from 100 DCSAGE models across all 7-day data input windows in our dataset. Each subplot represents sensitivity scores for one geographical region.

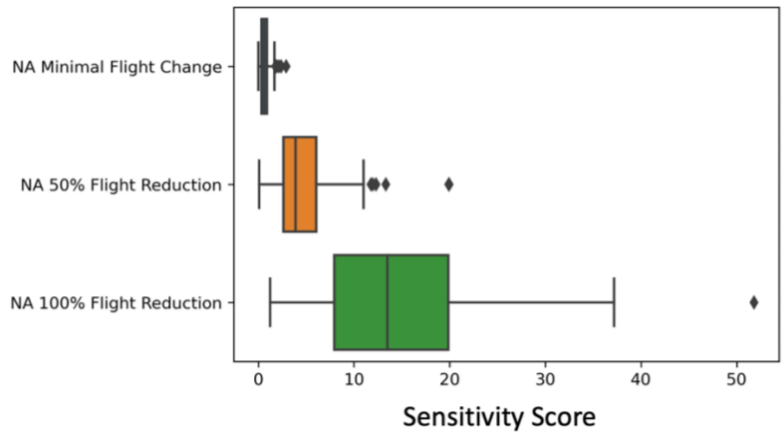

**Figure S6. DCSAGE Perturbation Steps.** Sensitivity score distributions shown through box plots when increasingly large flight reductions are applied to North America on a random test set window.

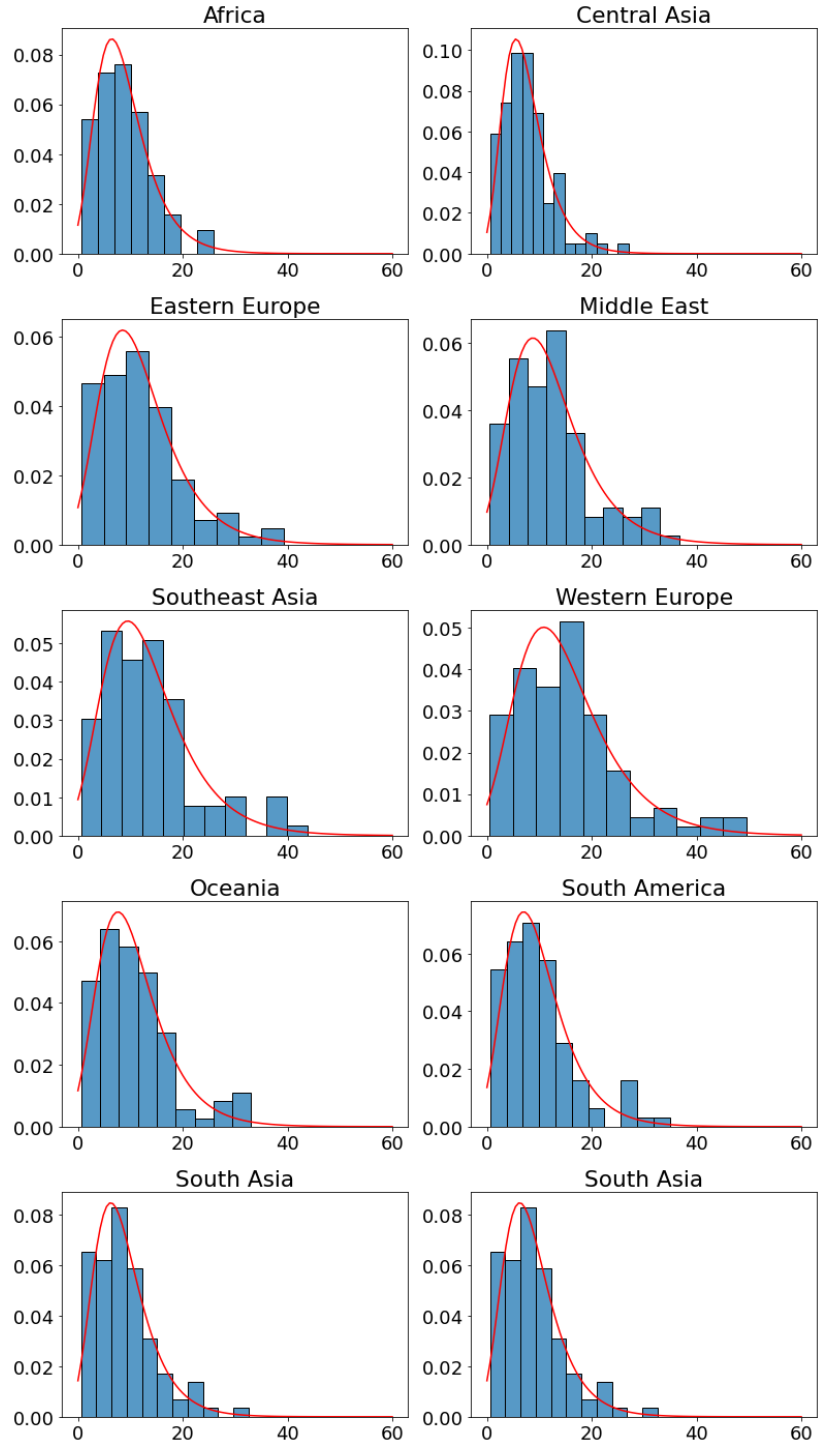

**Figure S7. DCSAGE Extreme Value distribution fits.** Gumbel distribution (Extreme Value Type I) distributions fitted on sensitivity score distributions for each geographical region on the first window of the dataset. Sensitivity score is plotted along the x-axis, and probability is shown on the y-axis. Sensitivity score distributions comprise of sensitivity scores for 100 DCSAGE models on the given window of the dataset.

## III. MPNN Results

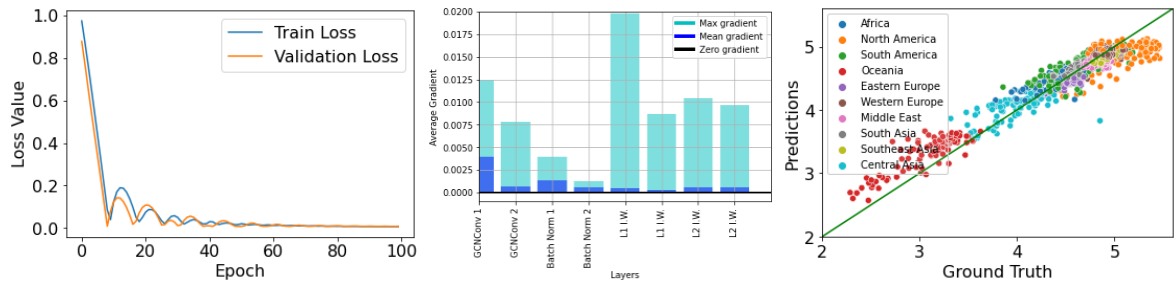

**Figure S8. MPNN Training Figures.** Loss curves (left), gradient flow plot (middle) and test set ground truth vs prediction correlation plot.

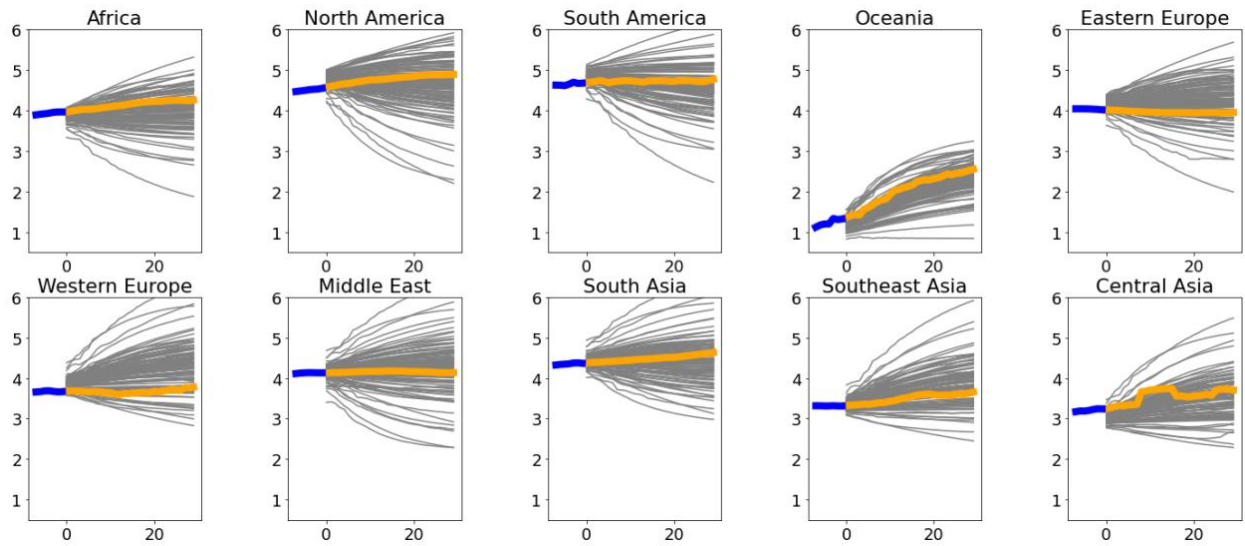

**Figure S9. MPNN Bias corrected predictions.** Recursive predictions of 100 MPNN models on the first dataset window after bias corrections.

## IV. Bias Correction Factors

**Table S1. Bias correction factors.** Bias correction factors calculated for DCSAGE and MPNN across the entire dataset.

|  | DCSAGE | MPNN |
|---|---|---|
| Africa | 0.970 | 0.969 |
| North America | 1.057 | 1.066 |
| South America | 1.048 | 1.064 |
| Oceania | 0.647 | 0.687 |
| Eastern Europe | 1.008 | 1.028 |
| Western Europe | 1.012 | 1.023 |
| Middle East | 1.007 | 1.003 |
| South Asia | 1.036 | 1.021 |
| Southeast Asia | 0.968 | 0.964 |
| Central Asia | 0.863 | 0.889 |